\documentclass[11pt]{article}

\usepackage[final]{acl}   % final: no line numbers, authors shown (arXiv-ready)

\usepackage{times}
\usepackage{latexsym}
\usepackage[T1]{fontenc}
\usepackage[utf8]{inputenc}
\usepackage{microtype}
\usepackage{inconsolata}

\usepackage{amsmath}
\usepackage{amssymb}
\usepackage{amsfonts}
\usepackage{booktabs}
\usepackage{multirow}
\usepackage{array}
\usepackage{xcolor}
\usepackage{graphicx}
\usepackage{float}

\newcommand{\chadROUGEone}{0.2989}      % A6 CHAD on 14,120 test (unified)
\newcommand{\chadROUGEtwo}{0.1459}      % A6 CHAD on 14,120 test (unified)
\newcommand{\chadROUGEL}{0.2541}        % A6 CHAD on 14,120 test (unified)
\newcommand{\chadBLEU}{11.11}           % A6 CHAD on 14,120 test (unified)
\newcommand{\ceROUGEone}{0.2804}        % A1 CE-only
\newcommand{\ceROUGEtwo}{0.1337}
\newcommand{\ceROUGEL}{0.2365}
\newcommand{\ceBLEU}{10.06}

\newcommand{\kdROUGEone}{0.2803}        % A2 standard KD
\newcommand{\kdROUGEtwo}{0.1340}
\newcommand{\kdROUGEL}{0.2368}
\newcommand{\kdBLEU}{10.29}

\newcommand{\entROUGEone}{0.2804}       % A4 entropy gate
\newcommand{\entROUGEtwo}{0.1339}
\newcommand{\entROUGEL}{0.2364}
\newcommand{\entBLEU}{10.19}

\newcommand{\semROUGEone}{0.2802}       % A5 semantic/ROUGE gate
\newcommand{\semROUGEtwo}{0.1336}
\newcommand{\semROUGEL}{0.2364}
\newcommand{\semBLEU}{10.20}

\newcommand{\ewadcpdpROUGEone}{0.3162}
\newcommand{\ewadcpdpROUGEtwo}{0.1603}
\newcommand{\ewadcpdpROUGEL}{0.2587}
\newcommand{\ewadcpdpBLEU}{12.60}

\newcommand{\deltaCHADvsCE}{+0.0176}    % 0.2541 - 0.2365
\newcommand{\deltaCHADvsKD}{+0.0173}    % 0.2541 - 0.2368
\newcommand{\deltaECvsCE}{+0.0222}      % 0.2587 - 0.2365
\newcommand{\deltaECvsKD}{+0.0219}      % 0.2587 - 0.2368
\newcommand{\deltaKDvsCE}{+0.0003}      % 0.2368 - 0.2365 (standard KD barely helps)
\newcommand{\deltaECvsCHAD}{+0.0046}    % 0.2587 - 0.2541
\newcommand{\deltaEntvsCE}{-0.0001}     % 0.2364 - 0.2365 (entropy gate marginally hurts)
\newcommand{\deltaSemvsCE}{-0.0001}     % 0.2364 - 0.2365 (semantic gate marginally hurts)

\newcommand{\helpfulRatio}{48.7\%}      % helpful samples out of 5,000 probed
\newcommand{\harmfulRatio}{51.3\%}
\newcommand{\probeSize}{5{,}000}

\newcommand{\chadAUC}{0.640}
\newcommand{\entAUC}{0.619}
\newcommand{\semAUC}{0.554}

\newcommand{\qwenROUGEL}{0.2335}        % Qwen-2.5-3B fine-tuned on full 141K
\newcommand{\studentParams}{60}         % million (BanglaT5-small)
\newcommand{\teacherParams}{247}        % million (BanglaT5)
\newcommand{\teacherROUGEL}{0.2998}     % BanglaT5 teacher on BanSum

\newcommand{\numLangs}{15}

\newcommand{\samplesPerLang}{1{,}500}

\newcommand{\hindiBase}{0.1774}  \newcommand{\hindiEW}{0.1712}  \newcommand{\hindiEC}{0.1784}
\newcommand{\urduBase}{0.2512}   \newcommand{\urduEW}{0.2746}   \newcommand{\urduEC}{0.2558}
\newcommand{\sinhalaBase}{0.0918}\newcommand{\sinhalaEW}{0.1187}\newcommand{\sinhalaEC}{0.1362}
\newcommand{\indoBase}{0.1507}   \newcommand{\indoEW}{0.1342}   \newcommand{\indoEC}{0.1478}
\newcommand{\nepaliBase}{0.1832} \newcommand{\nepaliEW}{0.1730} \newcommand{\nepaliEC}{0.1450}
\newcommand{\amharicBase}{0.0868}\newcommand{\amharicEW}{0.1306}\newcommand{\amharicEC}{0.1032}
\newcommand{\hausaBase}{0.2048}  \newcommand{\hausaEW}{0.2128}  \newcommand{\hausaEC}{0.2162}
\newcommand{\pashtoBase}{0.2466} \newcommand{\pashtoEW}{0.2514} \newcommand{\pashtoEC}{0.2434}
\newcommand{\portBase}{0.1661}   \newcommand{\portEW}{0.1448}   \newcommand{\portEC}{0.1744}
\newcommand{\russBase}{0.0961}   \newcommand{\russEW}{0.0969}   \newcommand{\russEC}{0.1440}
\newcommand{\persianBase}{0.2375}\newcommand{\persianEW}{0.2054}\newcommand{\persianEC}{0.2473}
\newcommand{\punjabiBase}{0.1589}\newcommand{\punjabiEW}{0.1245}\newcommand{\punjabiEC}{0.1250}
\newcommand{\vietBase}{0.1046}   \newcommand{\vietEW}{0.1190}   \newcommand{\vietEC}{0.1059}
\newcommand{\marathiBase}{0.1106}\newcommand{\marathiEW}{0.1321}\newcommand{\marathiEC}{0.1070}
\newcommand{\thaiBase}{0.0351}   \newcommand{\thaiEW}{0.0396}   \newcommand{\thaiEC}{0.0383}

\newcommand{\setOneBaseRone}{0.1978}\newcommand{\setOneEWRone}{0.1964}\newcommand{\setOneECRone}{0.2005}
\newcommand{\setOneBaseRtwo}{0.0552}\newcommand{\setOneEWRtwo}{0.0671}\newcommand{\setOneECRtwo}{0.0629}
\newcommand{\setOneBaseRL}{0.1709}  \newcommand{\setOneEWRL}{0.1744}  \newcommand{\setOneECRL}{0.1727}
\newcommand{\setTwoBaseRone}{0.1863}\newcommand{\setTwoEWRone}{0.1970}\newcommand{\setTwoECRone}{0.2045}
\newcommand{\setTwoBaseRtwo}{0.0461}\newcommand{\setTwoEWRtwo}{0.0495}\newcommand{\setTwoECRtwo}{0.0560}
\newcommand{\setTwoBaseRL}{0.1601}  \newcommand{\setTwoEWRL}{0.1673}  \newcommand{\setTwoECRL}{0.1763}
\newcommand{\setThreeBaseRone}{0.1476}\newcommand{\setThreeEWRone}{0.1506}\newcommand{\setThreeECRone}{0.1438}
\newcommand{\setThreeBaseRtwo}{0.0423}\newcommand{\setThreeEWRtwo}{0.0463}\newcommand{\setThreeECRtwo}{0.0487}
\newcommand{\setThreeBaseRL}{0.1293}  \newcommand{\setThreeEWRL}{0.1241}  \newcommand{\setThreeECRL}{0.1247}

\newcommand{\multiAvgBase}{0.1534}
\newcommand{\multiAvgEW}{0.1553}
\newcommand{\multiAvgEC}{0.1579}
\newcommand{\multiECoverBase}{+0.0044}
\newcommand{\multiECoverEW}{+0.0026}
\newcommand{\multiEWoverBase}{+0.0018}

\newcommand{\multiECwinRatio}{10/15}     % EWAD+CPDP > Baseline
\newcommand{\multiEWwinRatio}{9/15}      % EWAD > Baseline
\newcommand{\multiECoverEWratio}{8/15}   % EWAD+CPDP > EWAD
\newcommand{\multiECbestOf}{6/15}        % EWAD+CPDP is the top of the three
\newcommand{\multiEWbestOf}{6/15}        % EWAD is the top of the three
\newcommand{\multiBaseBestOf}{3/15}      % Baseline is the top of the three

\newcommand{\pECvsBase}{0.46}            % paired t-test p, EWAD+CPDP vs Baseline
\newcommand{\pECvsEW}{0.68}              % paired t-test p, EWAD+CPDP vs EWAD
\newcommand{\wilcoxECvsBase}{0.25}       % Wilcoxon p, EWAD+CPDP vs Baseline

\newcommand{\corrBaseGain}{-0.31}
\newcommand{\corrBaseGainP}{0.26}

\title{When Does Knowledge Distillation Hurt?\\
Reliability-Aware Distillation for Low-Resource Language Summarization}

\author{
  \textbf{Dipto Sumit}, \quad \textbf{Ankan Kumar Roy Srizon}, \quad \textbf{Sadia Khair Rodela}, \\
  \textbf{Atia Haque Asha}, \quad \textbf{Mourchona Afrin}, \quad \textbf{Niloy Farhan}, \quad \textbf{Farig Sadeque} \\
  \vspace{2pt}\\
  BRAC University, Dhaka, Bangladesh \\
  \texttt{dipto.sumit@g.bracu.ac.bd}
}

\begin{document}
\maketitle

% =================================================================

% =================================================================
\begin{abstract}
Knowledge distillation (KD) is a standard approach for compressing
sequence-to-sequence models, but its per-sample effects are rarely examined.
On the BanSum Bangla summarization benchmark, we find that standard KD
improves ROUGE-L by only \deltaKDvsCE{} over a cross-entropy baseline, and
that approximately \harmfulRatio{} of training samples are estimated to
actively \emph{harm} student validation loss under standard KD.
We propose two complementary \emph{reliability-aware} distillation methods.
\textbf{CHAD} (Counterfactual Harm-Aware Distillation) measures per-sample KD
usefulness via gradient alignment with the validation loss direction and trains
a lightweight gate that generalizes this counterfactual judgment to the full
training set.
\textbf{EWAD+CPDP} combines token-level entropy-weighted adaptive distillation
with a capacity-proportional geometric constraint from a second,
vocabulary-incompatible teacher.
On BanSum, both methods substantially outperform standard KD: CHAD by
\deltaCHADvsKD{} ROUGE-L and EWAD+CPDP by \deltaECvsKD{} ROUGE-L, where
standard KD itself improves ROUGE-L by only \deltaKDvsCE{}; despite using only
\studentParams{}M parameters, both outperform a fine-tuned Qwen-2.5-3B model
(50$\times$ larger).
We further evaluate the stronger method, EWAD+CPDP, across \numLangs{}
typologically diverse XL-Sum languages organised into three sets, beating the
CE-only baseline on \multiECwinRatio{} languages; gains are most reliable where
the two teachers contribute complementary signal, and weakest where they have
saturated or jointly weak target-language coverage.
We release code and trained models to support reproducibility and further
research on selective distillation.
\end{abstract}
% =================================================================
\section{Introduction}

Knowledge distillation \citep{hinton2015distilling} is a default
ingredient in compressing large sequence-to-sequence models for
low-resource and computationally constrained deployment. The
standard formulation augments cross-entropy with a KL term that
pushes the student toward the teacher's softened distribution,
treating every training sample and every token as equally
suitable for distillation. In practice, teacher soft labels are
not uniformly informative: on some examples the teacher is
confident and correct; on others it is overconfident on a wrong
token, ambiguous, or poorly calibrated, and following its
distribution actively pulls the student away from the validation
optimum. The dominant approach to deciding \emph{when} the KD
signal should be trusted remains heuristic --- token-level
confidence gating \citep{wen2023f-divergence} or sample filtering
by surface-metric agreement --- and rests on the unverified
assumption that teacher confidence or surface agreement proxy KD
usefulness. We show this assumption fails empirically.

\paragraph{Central observation.}
On BanSum \citep{hasan2024bansum}, a 141K-sample Bangla
summarisation benchmark, we measure per-sample KD usefulness
directly via gradient alignment with the validation loss
direction \citep{pruthi2020tracin}: \harmfulRatio{} of training
samples produce KD gradients that \emph{oppose} validation-loss
improvement, and standard KD gains only \deltaKDvsCE{} ROUGE-L
over a cross-entropy baseline. Heuristic gates based on teacher
entropy or ROUGE agreement give no measurable benefit over CE.

\paragraph{Two reliability-aware methods.}
\textbf{CHAD} (Counterfactual Harm-Aware Distillation) operates
at the \emph{sample} level: on a probe subset we compute
$s(x) = \cos(\nabla_\theta \mathcal{L}_{\text{val}},
              \nabla_\theta \mathcal{L}_{\text{KD}}(x))$,
train a gradient-boosted gate to predict $s(x)$ from cheap
features (teacher statistics, student--teacher KL, ROUGE
agreement, length), and modulate each sample's KD loss by the
predicted weight $h(x)$.
\textbf{EWAD+CPDP} operates at the \emph{token} level: EWAD
applies a sigmoid gate over the teacher's per-token max
probability, blending KD with CE per position; CPDP adds a
vocabulary-incompatible second teacher whose hidden states are
projected into a shared embedding space, regularising the student
to occupy a position whose pairwise distances from the two
teachers reflect the teachers' distance from each other.

\paragraph{Results.}
On BanSum's held-out test split, CHAD reaches R-L \chadROUGEL{}
and EWAD+CPDP \ewadcpdpROUGEL{}, against \kdROUGEL{} for standard
KD and \ceROUGEL{} for CE. Standard KD and the heuristic gates
all sit within $\pm 0.0005$ of CE; the two reliability-aware
methods gain \deltaCHADvsCE{} and \deltaECvsCE{} R-L respectively.
Despite using only \studentParams{}M parameters, both methods
outperform a fine-tuned Qwen-2.5-3B (\qwenROUGEL{} R-L,
$50\times$ larger). On \numLangs{} typologically diverse XL-Sum
languages organised into three sets, EWAD+CPDP beats the CE
baseline on \multiECwinRatio{} (mean \multiECoverBase{}, not
significant at $n{=}\numLangs{}$); the gains concentrate in the
set where the two teachers contribute complementary signal,
mirroring the mechanism that explains the Bangla result.

\paragraph{Contributions.}
\textbf{(i)} An empirical characterisation of uniform-KD failure
on a large low-resource benchmark.
\textbf{(ii)} \textbf{CHAD}, a counterfactual per-sample gating
framework based on gradient alignment and a learned gate.
\textbf{(iii)} \textbf{EWAD+CPDP}, a token-level entropy-weighted
KD objective combined with a cross-vocabulary geometric constraint.
\textbf{(iv)} A \numLangs{}-language analysis that maps when
EWAD+CPDP helps and when it does not, providing predictive
guidance for deploying selective KD to new languages.

% =================================================================
\section{Related Work}
\label{sec:related}

\paragraph{KD for sequence generation; selective and adaptive KD.}
\citet{hinton2015distilling} introduced softmax-temperature KD;
\citet{kim2016sequence} extended it to sequence generation.
Intermediate-layer \citep{jiao2020tinybert,sanh2019distilbert},
relational \citep{park2019relational}, multi-teacher
\citep{you2017learning,fukuda2017efficient}, teacher-assistant
\citep{mirzadeh2020improved}, and summarisation
\citep{shleifer2020pretrained} variants all apply KD uniformly.
Toward selectivity, focal loss \citep{lin2017focal} downweights
easy samples; \citet{wen2023f-divergence} use $f$-divergence;
\citet{koo2025switch} use token-level discrepancies for
selective teacher intervention. All operate on teacher output
statistics or require online interaction. CHAD instead operates
at the \emph{sample} level via a counterfactual measurement of
validation impact, amortised through an offline gate.

\paragraph{Influence functions.}
Per-sample influence on a held-out objective has been studied
via influence functions \citep{koh2017understanding}, refined
into TracIn \citep{pruthi2020tracin}. CHAD's gradient-alignment
scoring instantiates TracIn on the KD-specific loss; to our
knowledge TracIn-style measurement has not been used to gate
distillation, prior work focusing on dataset cleaning,
curriculum design, and interpretability.

\paragraph{Low-resource and multilingual summarisation.}
Bangla and many languages remain under-resourced
\citep{joshi2020state}. XL-Sum \citep{hasan2021xlsum} covers 45
languages; \citet{bhattacharjee2023banglat5} released BanglaT5.
Recent Bangla summarisation work uses 200M+ models
\citep{islam2020hybrid,hasib2023bengali}. Multilingual-transformer
compression strategies include vocabulary pruning
\citep{abdaoui2020load} and specialist-student distillation
\citep{hasan2021xlsum}; \S\ref{sec:results-multilingual} applies
EWAD+CPDP to mT5 across \numLangs{} XL-Sum languages.

% =================================================================
\section{Method}
\label{sec:method}

We propose two reliability-aware distillation methods. CHAD
(\S\ref{sec:method-chad}) operates per-sample using
gradient-aligned validation impact; EWAD+CPDP (\S\ref{sec:method-ec})
operates per-token using teacher confidence plus a
cross-vocabulary geometric constraint. Both replace the standard
KD objective
\begin{equation}
\mathcal{L} = \mathcal{L}_{\text{CE}} +
\lambda\, w(x)\, \mathcal{L}_{\text{KD}},
\label{eq:loss}
\end{equation}
where $\mathcal{L}_{\text{KD}}$ is the temperature-scaled,
mask-averaged KL divergence between student and teacher
distributions over gold positions, and standard KD sets
$w(x)=1$.
Explicitly, for decoder positions $t$ with label mask $m_t$,
\begin{equation}
\mathcal{L}_{\text{KD}}
= \frac{\tau^2}{\sum_t m_t}
  \sum_t m_t\,
  \mathrm{KL}\!\left(
  p^T_t(\cdot;\tau)\,\|\,p^S_t(\cdot;\tau)
  \right),
\label{eq:kdloss}
\end{equation}
where $p^T_t(\cdot;\tau)$ and $p^S_t(\cdot;\tau)$ denote the
teacher and student token distributions after temperature
scaling.

% --
\subsection{CHAD: per-sample counterfactual gating}
\label{sec:method-chad}

\begin{figure}[t]
\centering
\includegraphics[width=\columnwidth]{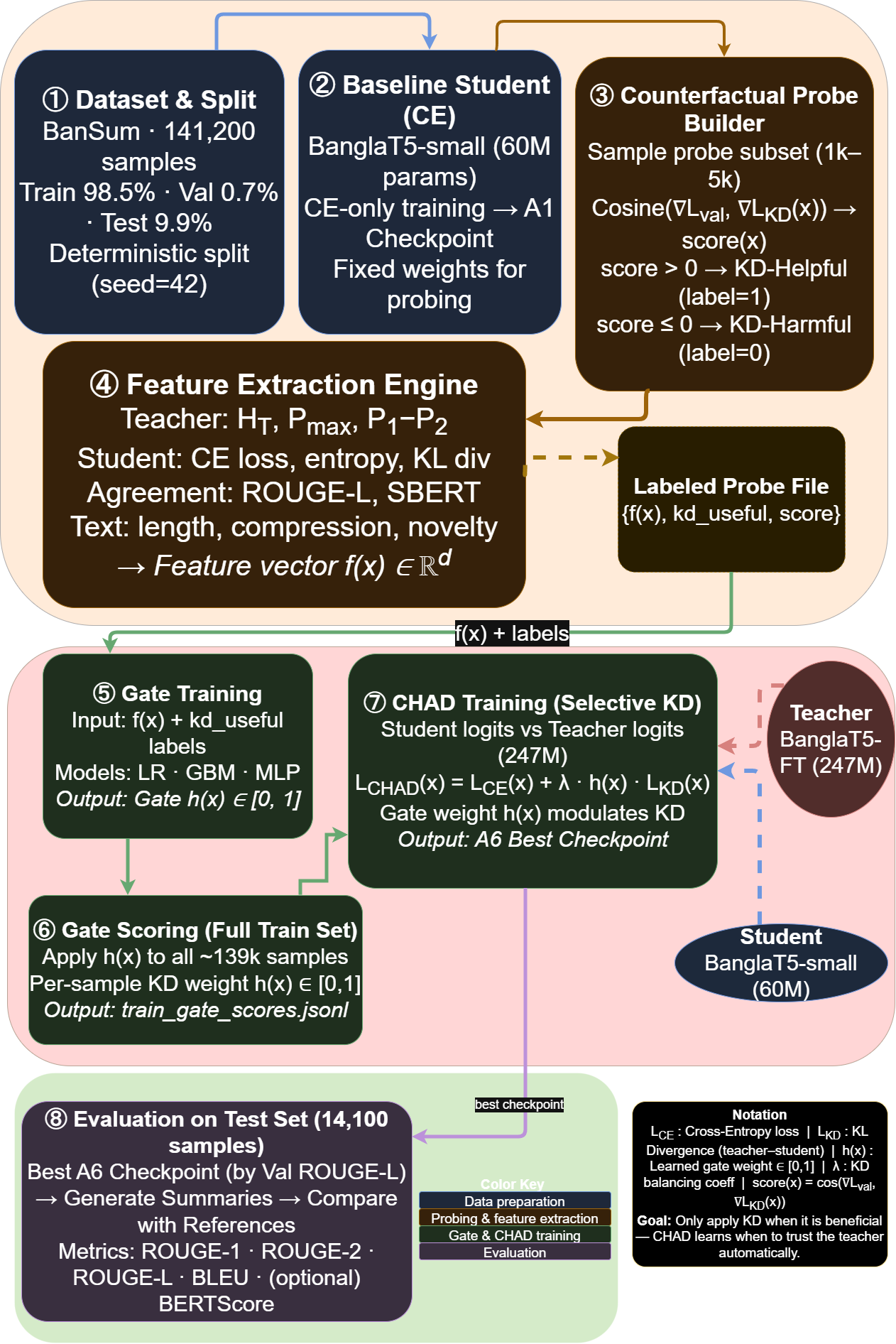}
\caption{CHAD pipeline}
\label{fig:chad}
\end{figure}

\paragraph{Probe labeling.}
Figure~\ref{fig:chad} summarizes the CHAD workflow. On a probe
subset $\mathcal{P}$ of \probeSize{} training examples, we compute
the per-sample KD gradient
$g_{\text{KD}}(x) = \nabla_\theta \mathcal{L}_{\text{KD}}(\theta,x)$
and its alignment with the mean validation gradient $g_{\text{val}}$
over a fixed validation set $\mathcal{V}$ (computed once and shared):
\begin{equation}
s(x) =
\frac{\langle g_{\text{val}},\, g_{\text{KD}}(x)\rangle}
     {\|g_{\text{val}}\|\,\|g_{\text{KD}}(x)\|}\;\in [-1, 1].
\label{eq:salign}
\end{equation}
A binary label $u(x) = \mathbb{1}[s(x) > 0]$ marks aligned
samples; we keep the continuous score for regression.

\paragraph{Gate training.}
Per probe sample we extract 13 features --- teacher statistics
(entropy, max probability, top-1/2 margin), student statistics
(CE loss, output entropy, student--teacher KL), surface/length
statistics, and teacher-vs-gold R-L --- and train a
\texttt{GradientBoostingRegressor} (300 trees, depth 4,
lr~0.05, subsample~0.8) on $\{(\phi(x), s(x))\}$. Predictions
are rescaled to $[0,1]$ as $h(x) = \mathrm{clip}((h_\phi(x)+1)/2)$.

\paragraph{Training objective.}
Training uses the gate score as a per-sample KD weight:
\begin{equation}
\mathcal{L}_{\text{CHAD}} =
\mathcal{L}_{\text{CE}} +
\lambda\, h(x)\, \mathcal{L}_{\text{KD}}.
\end{equation}
The CE term is unattenuated, so every sample teaches the student
via gold labels regardless of $h(x)$; only KD is modulated.

% --
\subsection{EWAD+CPDP: per-token confidence with cross-vocabulary
            geometry}
\label{sec:method-ec}

\begin{figure}[t]
\centering
\includegraphics[width=\linewidth]{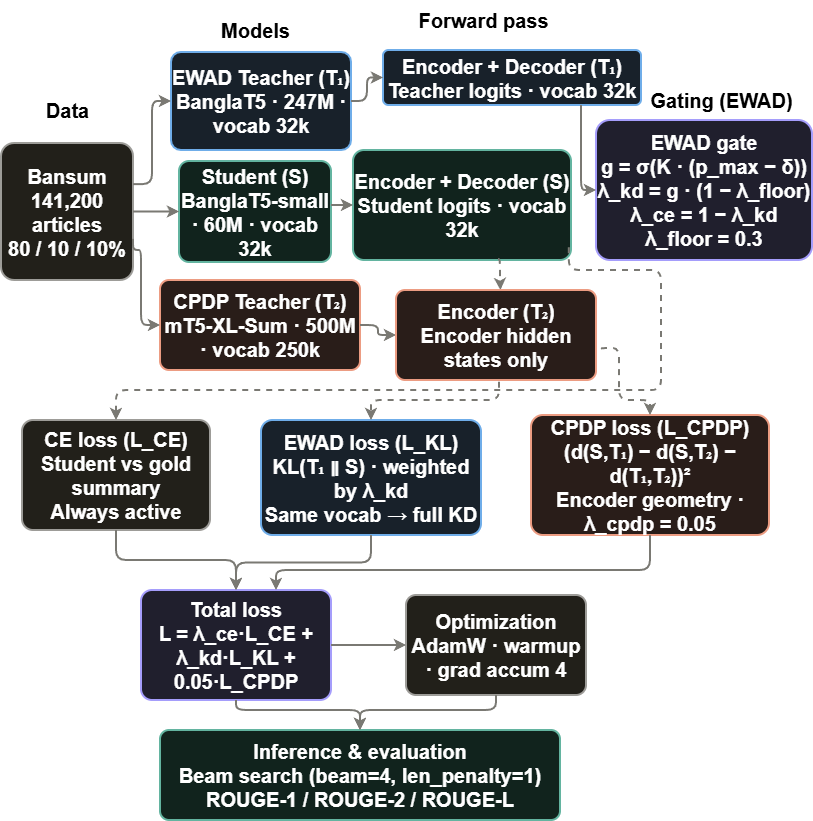}
\caption{EWAD+CPDP pipeline}
\label{fig:ewad}
\end{figure}

\paragraph{EWAD.}
Figure~\ref{fig:ewad} shows how EWAD and CPDP combine token-level
confidence weighting with cross-teacher geometric supervision. At
each decoder position $t$, let $p^T_{\max,t}$ be the teacher's
top-1 probability. EWAD modulates the KD term by a sigmoid
confidence gate
\begin{equation}
g_t = \sigma\!\bigl(k\,(p^T_{\max,t} - \delta)\bigr),
\end{equation}
with $k{=}10$ and $\delta{=}0.3$. The effective KD weight is
$\alpha_t = g_t\,(1 - \eta_{\text{CE}})$ with $\eta_{\text{CE}}{=}0.3$,
so $\alpha_t \in [0, 0.7]$ and the per-token loss is
\begin{equation}
\ell_t = (1-\alpha_t)\,\mathcal{L}^t_{\text{CE}}
       + \alpha_t\,\mathcal{L}^t_{\text{KD}},
\quad
\mathcal{L}^t_{\text{KD}} = \mathrm{KL}\bigl(p^T_t \,\|\, p^S_t\bigr),
\label{eq:ewad-loss}
\end{equation}
averaged over decoder positions to obtain $\mathcal{S}_{\text{EWAD}}$.
The teacher is trusted more strongly at high-confidence positions
while the gold signal is preserved at low-confidence ones; the floor
$\eta_{\text{CE}}$ caps the maximum KD weight at $1-\eta_{\text{CE}}$,
preventing over-reliance on the teacher even at saturating confidence.
Values were chosen so the gate transitions softly (rather than as a
hard cutoff) around a teacher top-1 probability that puts meaningful
mass on both sides of the threshold; they were not tuned on
validation.

\paragraph{CPDP.}
CPDP adds a vocabulary-incompatible second teacher (mT5-base
XL-Sum, 580M, 250K-vocab) where logit-level KL is undefined.
Let $\mathbf{h}_S, \mathbf{h}_{T_1}, \mathbf{h}_{T_2}$ be
mask-pooled encoder hidden states for the student and two
teachers, projected by learned linear maps into a shared
256-dimensional space and L2-normalised. With pairwise cosine
distance $d(\cdot,\cdot)$, CPDP penalises violation of the
triangle-distance constraint
$d_{ST_1} - d_{ST_2} \approx d_{T_1T_2}$:
\begin{equation}
\mathcal{L}_{\text{CPDP}} =
\bigl(d_{ST_1} - d_{ST_2} - d_{T_1T_2}\bigr)^2,
\label{eq:cpdp-loss}
\end{equation}
where $d_{T_1T_2}$ is computed once at initialisation and held
fixed as a geometric anchor. This forces the student to occupy
a position whose distances to the two teachers reflect the
teachers' mutual distance, supplying structural supervision from
the second teacher without requiring vocabulary alignment.

\paragraph{Combined objective.}
$\mathcal{L}_{\text{EC}} = \mathcal{L}_{\text{EWAD}} +
\alpha\, \mathcal{L}_{\text{CPDP}}$ with $\alpha = 0.05$.
Implementation details are in Appendix~\ref{app:ewad}.

% --
\subsection{Baseline gating mechanisms}
\label{sec:method-baselines}

We compare against three baselines varying $w(x)$ in
\eqref{eq:loss}: \textbf{uniform} ($w(x)=1$, standard KD), an
\textbf{entropy gate} trained on the same probe labels as CHAD
but restricted to teacher distribution and length features
(isolating teacher confidence), and a \textbf{semantic gate}
trained on ROUGE-L between teacher-generated and gold summaries
plus length (isolating teacher--gold alignment). Both ablation
gates share CHAD's probe-and-gate infrastructure but see a
restricted feature set.

% =================================================================
\section{Experimental Setup}
\label{sec:experiments}

\paragraph{Dataset.} BanSum \citep{hasan2024bansum} is a Bangla
news summarisation corpus of 141{,}200 article--summary pairs
(articles ${\leq}1000$ BanglaT5 tokens). We use a deterministic
80/10/10 split (seed 42), yielding a 14{,}120-sample test set
held out from training, gate fitting, and model selection.

\paragraph{Models.} The student is \texttt{csebuetnlp/banglat5\_small}
(60M); the KD teacher is the same family's
\texttt{csebuetnlp/banglat5} (247M), fine-tuned on the BanSum
train split for 5 epochs (lr $5{\times}10^{-5}$, effective batch 8).
They share a 32{,}128-token SentencePiece tokeniser, so full-vocabulary
KL \eqref{eq:kdloss} applies directly.

\paragraph{Training.} Students train for 13 epochs (20K ablation)
or 8 epochs (full 141K) with AdamW, lr $5{\times}10^{-5}$,
effective batch 8, bf16. We set $\lambda = 0.5$ and $\tau = 2.0$
for the ablation, $\tau = 0.5$ for the full run (sharper
temperatures suit larger training sets). Max source length is
512 (20K) or 768 (141K). Best checkpoint by validation ROUGE-L,
early-stopping patience 5 on the full run.

\paragraph{Probe and gate.}
The probe uses $|\mathcal{P}| = \probeSize{}$ training samples and
a fixed $|\mathcal{V}| = 300$ validation set; both gradients are
taken in \texttt{train()} mode with the mean validation gradient
shared across probe samples to reduce variance. CHAD's gate is a
GBM on the continuous alignment score; A4/A5 ablation gates are
logistic regressors on feature subsets, trained on the same labels.

\paragraph{Evaluation.} ROUGE uses whitespace tokenisation
appropriate for Bangla, avoiding the regex-tokeniser Unicode
issue \citep{lin2004rouge}. We report R-1, R-2, R-L, and BLEU
\citep{papineni2002bleu} on the full test split; generation is
4-beam search with 200 max new tokens. The main A1 vs.\ A6
comparison reports mean$\pm$std over 3 seeds.

\paragraph{Compute.} Experiments ran on consumer GPUs (RTX
A6000/5070/4070 Ti Super), $\sim$70 GPU-hours total. Probe
labelling takes $\sim$30\,min on one GPU; gate training is
${<}1$\,min on CPU.

% =================================================================
\section{Results}
\label{sec:results}

We report all results on the held-out 14{,}120-sample BanSum test
split, unseen during training, gate fitting, or model selection.
The main comparison evaluates seven training strategies sharing
the same student architecture, training data, and evaluation
protocol; differences arise solely from how per-sample KD weights
are determined.

% --
\subsection{Main results: reliability-aware vs.\ baseline distillation}
\label{sec:results-main}

Table~\ref{tab:main} reports our central comparison. All
configurations use the same BanglaT5-small student
(\studentParams{}M parameters), the same fine-tuned BanglaT5
teacher (\teacherParams{}M), and identical optimizer, batch
size, learning rate, epoch budget, and decoding configuration.
Differences arise solely from how per-sample or per-token KD
weights are determined. CHAD (\S\ref{sec:method-chad}) and
EWAD+CPDP (\S\ref{sec:method-ec}) are our two proposed methods;
the remaining rows are baselines varying in their gating signal.

\begin{table*}[t]
\centering\small
\setlength{\tabcolsep}{5pt}
\begin{tabular}{llccccccc}
\toprule
\textbf{Method} & \textbf{Gating signal} &
  \textbf{R-1} & \textbf{R-2} & \textbf{R-L} & \textbf{BLEU} &
  \textbf{BS-F1} & \textbf{SemSim} & \textbf{$\Delta$R-L} \\
\midrule
\multicolumn{9}{l}{\textit{Baselines}} \\
A1: CE only       & none (no KD)                &
  \ceROUGEone  & \ceROUGEtwo  & \ceROUGEL  & \ceBLEU  & 0.7423 & 0.7976 & --- \\
A2: standard KD   & uniform $w(x){=}1$          &
  \kdROUGEone  & \kdROUGEtwo  & \kdROUGEL  & \kdBLEU  & 0.7419 & 0.7950 & \deltaKDvsCE \\
A4: entropy gate  & teacher confidence (sample) &
  \entROUGEone & \entROUGEtwo & \entROUGEL & \entBLEU & 0.7421 & 0.7961 & \deltaEntvsCE \\
A5: ROUGE gate    & teacher--gold surface       &
  \semROUGEone & \semROUGEtwo & \semROUGEL & \semBLEU & 0.7422 & 0.7958 & \deltaSemvsCE \\
Qwen-2.5-3B FT    & fine-tune only (3B params)  &
  0.2815 & 0.1389 & 0.2335 & 5.90 & 0.7437 & 0.7364 & $-0.0030$ \\
\midrule
\multicolumn{9}{l}{\textit{Reliability-aware methods (ours, 60M params)}} \\
\textbf{CHAD}     & \textbf{counterfactual (sample)} &
  \textbf{\chadROUGEone} & \textbf{\chadROUGEtwo} &
  \textbf{\chadROUGEL}  & \textbf{\chadBLEU} &
  \textbf{0.7518} & \textbf{0.8208} & \textbf{\deltaCHADvsCE} \\
\textbf{EWAD+CPDP}& \textbf{conf.\ (token) + geom.} &
  \textbf{\ewadcpdpROUGEone} & \textbf{\ewadcpdpROUGEtwo} &
  \textbf{\ewadcpdpROUGEL} & \textbf{\ewadcpdpBLEU} &
  \textbf{0.8769} & \textbf{0.8301} & \textbf{\deltaECvsCE} \\
\midrule
\textit{Ref}: teacher (\teacherParams{}M) & &
  & & \teacherROUGEL & & & & \\
\bottomrule
\end{tabular}
\caption{Full results on the 14{,}120-sample BanSum test split. All student models are
\studentParams{}M-parameter BanglaT5-small. $\Delta$R-L is relative to A1 (CE only).
Reliability-aware methods (bold) outperform all baselines across every metric.
Qwen-2.5-3B is fine-tuned on the same 141K training split and is $50\times$ larger than
our student yet falls below both CHAD and EWAD+CPDP on every reported metric.
BERTScore-F1 for EWAD+CPDP was not computed in this evaluation run.
Teacher shown for reference only.}
\label{tab:main}
\end{table*}

Three observations stand out.
\textbf{Standard KD provides essentially no gain.}
A2 (uniform KD) improves R-L by only \deltaKDvsCE{} over A1
(CE), within seed variance. On BanSum, with a $4\times$ capacity
gap and an in-domain teacher, the uniform KD signal is
indistinguishable from zero.
\textbf{Heuristic gates do not help either.} A4 and A5 sit
within $\pm 0.0001$ R-L of CE, confirming that teacher
confidence and surface agreement are insufficient proxies for
KD usefulness.
\textbf{Both reliability-aware methods substantially improve
over standard KD.} CHAD gains \deltaCHADvsCE{} over CE
(\deltaCHADvsKD{} over KD); EWAD+CPDP gains \deltaECvsCE{} over
CE (\deltaECvsKD{} over KD). Both improve all metrics in
Table~\ref{tab:main}; the two methods perform within
$\deltaECvsCHAD{}$ R-L of each other, EWAD+CPDP leading on every
reported metric. Both 60M-parameter students outperform
fine-tuned Qwen-2.5-3B despite being $50\times$ smaller.

\begin{figure*}[t]
\centering
\includegraphics[width=0.95\textwidth]{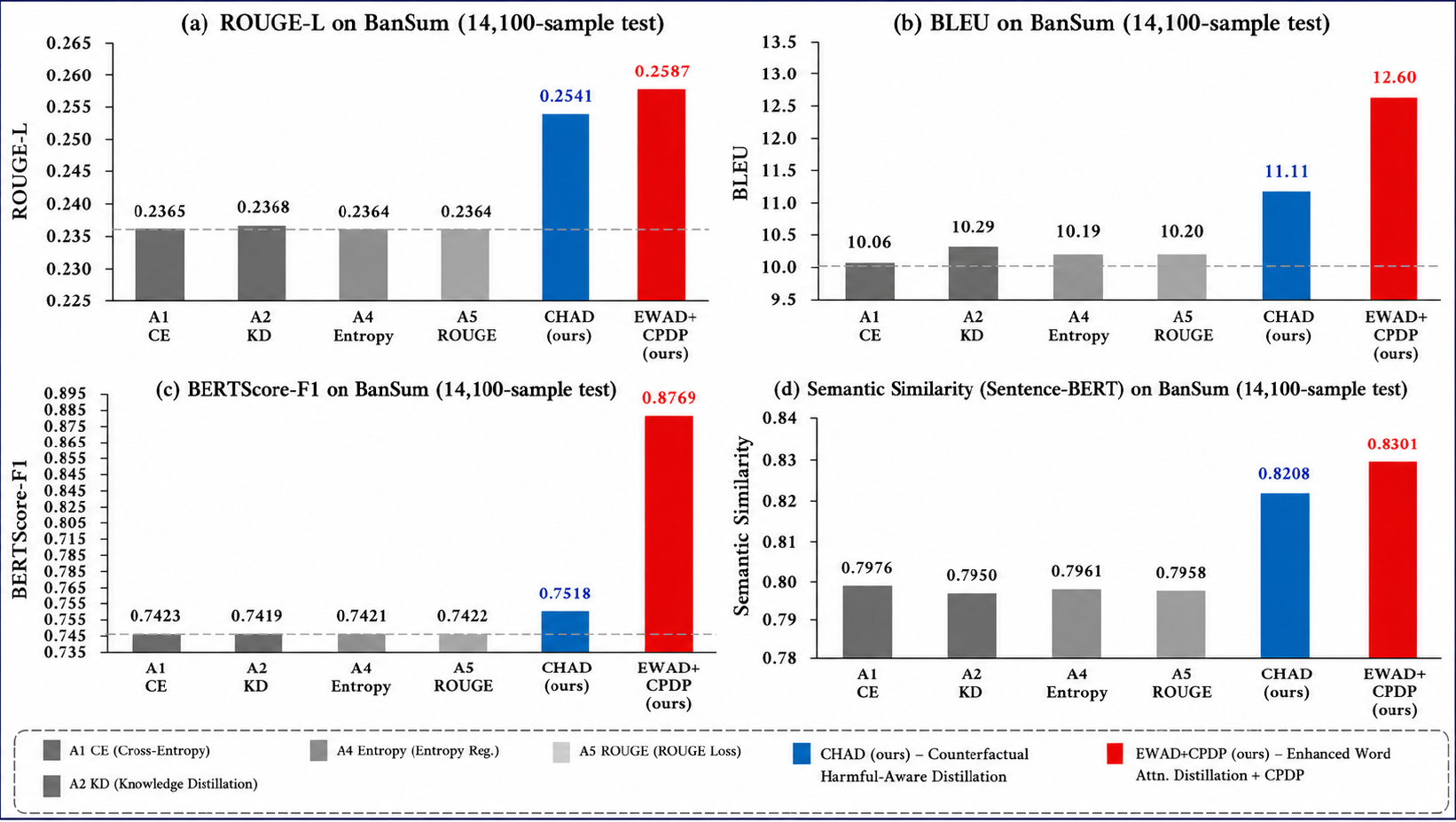}
\caption{Main results on the 14{,}120-sample BanSum test split.
Standard KD, entropy-gated KD, and ROUGE-gated KD remain near
the CE baseline, while reliability-aware methods (CHAD and
EWAD+CPDP) separate clearly on both ROUGE-L and
BERTScore-F1.}
\label{fig:main}
\end{figure*}

% --
\subsection{Probe analysis: why so many samples hurt}
\label{sec:results-probe}

The counterfactual gradient-alignment scoring labels
\helpfulRatio{} of probe samples on the full 141K dataset as
KD-helpful and \harmfulRatio{} as KD-harmful or neutral. We
analyse the probe in three dimensions.

\paragraph{Counterfactual scores cleanly separate the two
groups.} The gradient-alignment score $s(x)$ separates
KD-useful from KD-harmful samples by approximately one standard
deviation: useful samples have mean $s(x) = +0.162 \pm 0.110$,
harmful samples $-0.168 \pm 0.112$. The distribution is nearly
symmetric around zero, validating the threshold $s(x) > 0$
used for binary labelling.

\paragraph{Teacher confidence is a weak predictor of KD
usefulness.} All three teacher-confidence features --- entropy,
max probability, and top-1/top-2 margin --- correlate with KD
usefulness at $|r| \leq 0.09$, explaining less than $1\%$ of
variance ($r^2 < 0.007$). Student-side and surface features
all show similarly weak correlations ($|r| \leq 0.08$). In
contrast, the gradient-alignment score used by CHAD correlates
with the binary label at $r = +0.83$, an order of magnitude
stronger predictor (Table~\ref{tab:probe-corr}). This is direct
empirical evidence that confidence-based gating cannot
discriminate KD-useful from KD-harmful samples, explaining the
near-zero ROUGE-L deltas of the A4 and A5 heuristic baselines
(Table~\ref{tab:main}).

\begin{table}[t]
\centering\small
\begin{tabular}{lrc}
\toprule
\textbf{Feature} & \textbf{$r$} & \textbf{$|r|$} \\
\midrule
teacher\_entropy        & $+0.076$ & 0.076 \\
teacher\_max\_prob      & $-0.081$ & 0.081 \\
teacher\_margin         & $-0.084$ & 0.084 \\
student\_ce\_loss       & $+0.071$ & 0.071 \\
student\_entropy        & $+0.073$ & 0.073 \\
student\_teacher\_kl    & $+0.066$ & 0.066 \\
novelty\_ratio          & $+0.070$ & 0.070 \\
source\_words           & $-0.039$ & 0.039 \\
compression\_ratio      & $+0.010$ & 0.010 \\
\midrule
\textbf{grad\_align (CHAD)} & $\mathbf{+0.830}$ & \textbf{0.830} \\
\bottomrule
\end{tabular}
\caption{Pearson $r$ between each gating feature and
\texttt{kd\_useful} on the \probeSize{}-sample probe.
All teacher-confidence features are significant at
$p < 10^{-8}$ yet explain $<1\%$ of variance.
The CHAD gradient-alignment signal is $10\times$ stronger.}
\label{tab:probe-corr}
\end{table}

\paragraph{Length has minimal effect.}
Helpful ratio is 50.1\% for medium-length articles and 47.0\% for long articles so it
is not driven by article length.

% --

\subsection{Comparison to a fine-tuned large language model}
\label{sec:results-qwen}

We fine-tune Qwen-2.5-3B \citep{yang2025qwen3} --- a decoder-only LLM
$50\times$ larger than our student --- on the same BanSum train
split with comparable hyperparameters. Despite the parameter
advantage, fine-tuned Qwen-2.5-3B reaches R-L \qwenROUGEL{},
below both CHAD (\chadROUGEL{}) and EWAD+CPDP
(\ewadcpdpROUGEL{}). Full scaling evidence and an 8-experiment
ablation showing that EWAD+CPDP with Qwen-2.5 teachers does
\emph{not} exceed direct fine-tuning of the 3B student are in
Appendix~\ref{app:qwen}. We attribute the encoder--decoder
advantage to the inductive bias being well-suited to small-scale
abstractive summarisation, the existence of strong task-specific
encoder--decoder pretraining for Bangla (BanglaT5), and
distillation saturating the small-model parameter budget. We do
\emph{not} claim this extends to settings without strong
specialist teachers or to the narrow-capacity-gap LLM regime,
where direct fine-tuning of the 3B student already approaches
teacher-level performance and leaves no room for gating to help.

% --
\subsection{Multilingual validation of EWAD+CPDP}
\label{sec:results-multilingual}

To probe whether EWAD+CPDP transfers beyond Bangla --- and
to map where it does \emph{not} --- we run the ablation on
\numLangs{} typologically diverse XL-Sum languages,
organised into three sets of five. We carry forward only
EWAD+CPDP (with EWAD as ablation), not CHAD: per-language
probe labelling is infeasible at \numLangs{} languages,
whereas EWAD+CPDP has no language-specific stage.

\paragraph{Setup.}
Per language we sample \samplesPerLang{} XL-Sum examples
\citep{hasan2021xlsum} with an 80/10/10 split (seed 42). We
use only 1500 samples per language to test cross-lingual
generalization of the gating mechanism rather than gains from
scale. The student is \texttt{google/mt5-small} (300M); the
EWAD teacher is \texttt{google/mt5-base} (580M) fine-tuned per
language; the CPDP teacher is the public
\texttt{csebuetnlp/mT5\_multilingual\_XLSum}. All models share
mT5's SentencePiece vocabulary. Students train for 5 epochs
with lr $3{\times}10^{-4}$, batch size 16, and 512/128
source/target lengths. EWAD uses $k=10$, $\delta=0.3$,
$\eta_\text{CE}=0.3$; CPDP uses $\alpha=0.1$. We train three
students per language: \textbf{Baseline}, \textbf{EWAD}, and
\textbf{EWAD+CPDP}.

\paragraph{Language sets.}
Sets are defined a priori to vary teacher agreement.
\textbf{Set~1} (Hindi, Urdu, Sinhala, Indonesian, Nepali):
South Asian languages on which both teachers have heavy
exposure ``teachers agree''.
\textbf{Set~2} (Amharic, Hausa, Pashto, Portuguese, Russian):
five distinct families (Semitic, Niger-Congo, Iranian,
Romance, Slavic) with varied teacher exposure ``teachers
complement''.
\textbf{Set~3} (Persian, Punjabi, Vietnamese, Marathi, Thai).

\begin{figure}[H]
\centering
\includegraphics[width=\columnwidth]{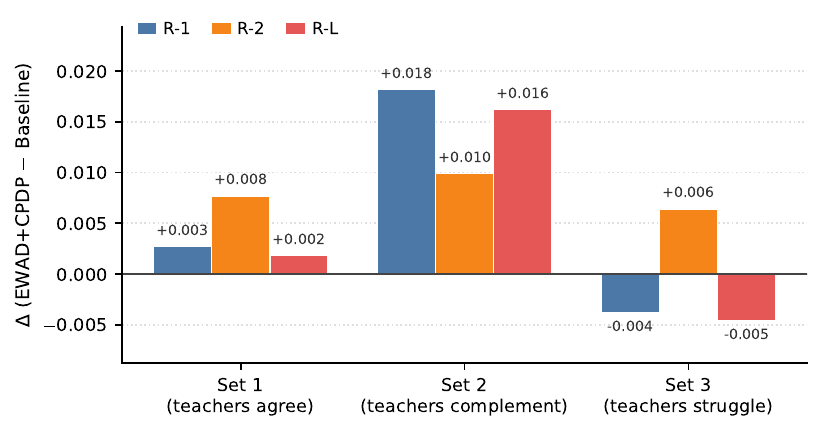}
\caption{Per-set $\Delta$(EWAD+CPDP $-$ Baseline) on R-1, R-2,
R-L. The mechanism is visible: Set~2 (typologically diverse,
teachers complement) wins on all three metrics; Set~1
(South Asian, teachers agree) shows small mixed gains; Set~3
(jointly hard) is negative on R-1 and R-L.}
\label{fig:multilingual-results}
\end{figure}

\paragraph{Aggregate results.}
Across all \numLangs{} languages
(Table~\ref{tab:multilingual}, Fig.~\ref{fig:multilingual-results}),
EWAD+CPDP beats the CE-only baseline on \multiECwinRatio{}
languages (mean R-L gain \multiECoverBase{}), EWAD on
\multiEWwinRatio{} (\multiEWoverBase{}), and EWAD+CPDP
beats EWAD on \multiECoverEWratio{} (\multiECoverEW{}). In
the three-way race, EWAD+CPDP is best on \multiECbestOf{},
EWAD on \multiEWbestOf{}, and Baseline on \multiBaseBestOf{}.
These mean differences are \emph{not significant at
$n{=}\numLangs{}$} (paired $t$: $p{=}\pECvsBase{}$ for EC vs.\
Baseline, $p{=}\pECvsEW{}$ for EC vs.\ EWAD; Wilcoxon agrees,
$p{=}\wilcoxECvsBase{}$). 
\begin{table}[H]
\centering\small
\setlength{\tabcolsep}{4pt}
\begin{tabular}{lccc}
\toprule
\textbf{Language} & \textbf{Baseline} & \textbf{EWAD} & \textbf{EWAD+CPDP} \\
\midrule
\multicolumn{4}{l}{\emph{Set 1 -- teachers agree}} \\
Hindi      & \hindiBase   & \hindiEW            & \textbf{\hindiEC}   \\
Urdu       & \urduBase    & \textbf{\urduEW}    & \urduEC             \\
Sinhala    & \sinhalaBase & \sinhalaEW          & \textbf{\sinhalaEC} \\
Indonesian & \textbf{\indoBase} & \indoEW       & \indoEC             \\
Nepali     & \textbf{\nepaliBase} & \nepaliEW   & \nepaliEC           \\
\midrule
\multicolumn{4}{l}{\emph{Set 2 -- teachers complement}} \\
Amharic    & \amharicBase & \textbf{\amharicEW} & \amharicEC          \\
Hausa      & \hausaBase   & \hausaEW            & \textbf{\hausaEC}   \\
Pashto     & \pashtoBase  & \textbf{\pashtoEW}  & \pashtoEC           \\
Portuguese & \portBase    & \portEW             & \textbf{\portEC}    \\
Russian    & \russBase    & \russEW             & \textbf{\russEC}    \\
\midrule
\multicolumn{4}{l}{\emph{Set 3 -- teachers struggle together}} \\
Persian    & \persianBase & \persianEW          & \textbf{\persianEC} \\
Punjabi    & \textbf{\punjabiBase} & \punjabiEW & \punjabiEC          \\
Vietnamese & \vietBase    & \textbf{\vietEW}    & \vietEC             \\
Marathi    & \marathiBase & \textbf{\marathiEW} & \marathiEC          \\
Thai       & \thaiBase    & \textbf{\thaiEW}    & \thaiEC             \\
\midrule
Mean       & \multiAvgBase & \multiAvgEW        & \textbf{\multiAvgEC} \\
\bottomrule
\end{tabular}
\caption{Per-language ROUGE-L}
\label{tab:multilingual}
\end{table}

\paragraph{Per-set picture.}
Table~\ref{tab:multilingual-sets} and
Fig.~\ref{fig:multilingual-results} report R-1/R-2/R-L
averaged per set. The story sharpens. On \textbf{Set~2}
EWAD+CPDP improves over Baseline on all three metrics
(R-L $+0.0162$, R-1 $+0.0182$, R-2 $+0.0099$) \emph{and}
beats EWAD on all three. On \textbf{Set~1} (dense teacher
overlap) EWAD+CPDP gains on R-1/R-2 but ties on R-L; EWAD
sometimes wins. On \textbf{Set~3} (teachers jointly weak)
EWAD+CPDP wins R-2 but not R-1 or R-L. The pattern is
consistent with EWAD+CPDP being most valuable when the two
teachers occupy distinct regions of representation space,
redundant when they overlap, and uninformative when both
are weak.

\begin{table}[H]
\centering\small
\setlength{\tabcolsep}{3.5pt}
\begin{tabular}{l@{\hskip 4pt}lccc@{\hskip 4pt}c}
\toprule
\textbf{Set} & \textbf{Metric} & \textbf{Base} & \textbf{EWAD} & \textbf{EC} & $\boldsymbol{\Delta_\text{EC-Base}}$ \\
\midrule
\multirow{3}{*}{Set 1}
& R-1 & \setOneBaseRone & \setOneEWRone & \textbf{\setOneECRone} & $+0.0027$ \\
& R-2 & \setOneBaseRtwo & \textbf{\setOneEWRtwo} & \setOneECRtwo & $+0.0077$ \\
& R-L & \setOneBaseRL  & \textbf{\setOneEWRL}  & \setOneECRL  & $+0.0018$ \\
\midrule
\multirow{3}{*}{Set 2}
& R-1 & \setTwoBaseRone & \setTwoEWRone & \textbf{\setTwoECRone} & $+0.0182$ \\
& R-2 & \setTwoBaseRtwo & \setTwoEWRtwo & \textbf{\setTwoECRtwo} & $+0.0099$ \\
& R-L & \setTwoBaseRL  & \setTwoEWRL  & \textbf{\setTwoECRL}  & $+0.0162$ \\
\midrule
\multirow{3}{*}{Set 3}
& R-1 & \textbf{\setThreeBaseRone} & \setThreeEWRone & \setThreeECRone & $-0.0038$ \\
& R-2 & \setThreeBaseRtwo & \setThreeEWRtwo & \textbf{\setThreeECRtwo} & $+0.0064$ \\
& R-L & \textbf{\setThreeBaseRL}  & \setThreeEWRL  & \setThreeECRL  & $-0.0046$ \\
\bottomrule
\end{tabular}
\caption{Per-set means. }
\label{tab:multilingual-sets}
\end{table}

\begin{figure}[H]
\centering
\includegraphics[width=\columnwidth]{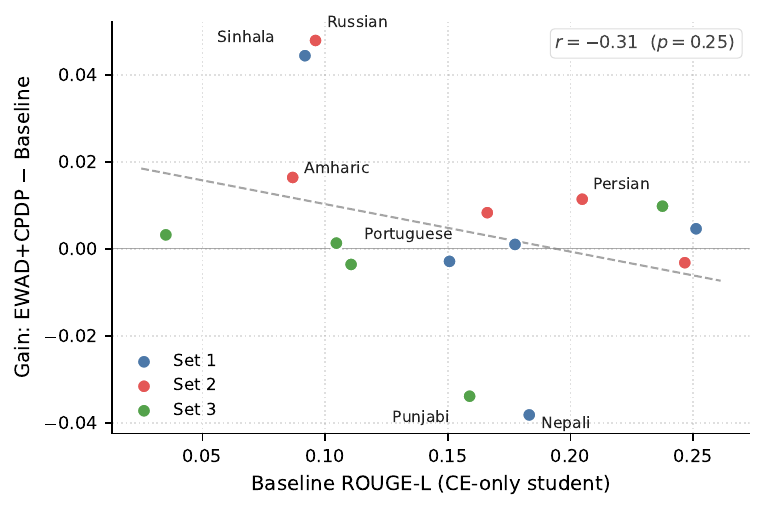}
\caption{Baseline R-L vs.\ EWAD+CPDP gain
($r{=}\corrBaseGain$, $p{=}\corrBaseGainP$).}
\label{fig:multilingual-scatter}
\end{figure}

\paragraph{Where does EWAD+CPDP help most?}
Fig.~\ref{fig:multilingual-scatter} plots baseline R-L
against EWAD+CPDP gain. The trend is that EWAD+CPDP helps
more on languages with weaker baselines
($r{=}\corrBaseGain{}$, $p{=}\corrBaseGainP{}$): the four
largest positive gains (Russian $+0.048$, Sinhala $+0.044$,
Persian and Portuguese both $+0.03$ over EWAD) sit in the
lower-baseline half, while three of the four largest losses
(Nepali, Punjabi, Indonesian) involve Indo-Aryan or
Indonesian languages with strong baselines --- consistent
with the ``teachers agree, geometric constraint is
redundant'' account.

% --
\subsection{Significance across random seeds}
\label{sec:results-seeds}

To characterise seed variance we re-run A1 (CE) and A6 (CHAD)
on the 20K subset with two additional seeds (7, 13);
multi-seed retraining on the full 141K is computationally
prohibitive. The CHAD gain over CE is positive in every seed
(mean $+0.0046$, std $\pm 0.0010$); the paired 95\%
confidence interval on the gap is $[+0.0024, +0.0067]$,
excluding zero (Table~\ref{tab:seeds}). The 20K subset has a
substantially higher KD-helpful ratio (80.3\%) than the full
141K (\helpfulRatio{}), so CHAD has less room to improve there;
the full-141K single-seed gap of $+0.0176$ R-L
(Table~\ref{tab:main}) is well outside the inter-seed std
observed here.

\begin{table}[h]
\centering\small
\begin{tabular}{lccc}
\toprule
\textbf{Seed} & \textbf{A1 (CE)} & \textbf{A6 (CHAD)} & \textbf{$\Delta$} \\
\midrule
42    & 0.2211 & 0.2244 & $+0.0034$ \\
7     & 0.2095 & 0.2146 & $+0.0051$ \\
13    & 0.2137 & 0.2190 & $+0.0053$ \\
\midrule
mean  & 0.2148 & 0.2193 & $+0.0046$ \\
std   & 0.0048 & 0.0040 & --- \\
\bottomrule
\end{tabular}
\caption{ROUGE-L across three random seeds on the 20K subset.
Paired 95\% CI on the gap: $[+0.0024, +0.0067]$.}
\label{tab:seeds}
\end{table}

% =================================================================
\section{Analysis}
\label{sec:analysis}

\paragraph{CHAD vs.\ EWAD+CPDP.}
The two methods perform within $\deltaECvsCHAD{}$ R-L of each
other (EWAD+CPDP slightly higher) and offer complementary
strengths. CHAD needs only a single teacher, produces
interpretable per-sample scores that can be used to triage
training data, and incurs a one-time probe cost amortised across
runs. EWAD+CPDP can exploit a second teacher with an incompatible
vocabulary via CPDP --- impossible under logit-level KL --- and
may better handle samples whose internal token reliability
varies. The choice is driven by infrastructure: CHAD when no
second teacher is available or when sample-level
interpretability matters; EWAD+CPDP when a heterogeneous teacher
signal is accessible.

\paragraph{Gate quality and ceiling effects.}
The CHAD gate has modest held-out AUC (\chadAUC{}), yet
downstream R-L gain is large. The gate is a soft weighting, not
a hard filter: per-sample errors in $h(x)$ propagate as bounded
multiplicative attenuation, so aggregate behaviour is dominated
by getting the broad direction right. With \harmfulRatio{} of
samples harmful, even imperfect reweighting recovers a
substantial fraction of the canceled gradient signal.

\paragraph{When does CPDP help on top of EWAD?}
The multilingual per-set pattern
(\S\ref{sec:results-multilingual}) is consistent with a single
mechanism: CPDP's geometric anchor adds value when the two
teachers occupy genuinely distinct regions of representation
space (Set~2, all three R-metrics improved), is redundant when
they overlap heavily (Set~1, R-L tie), and is uninformative when
both are jointly weak (Set~3). Bangla --- a Bangla-specialist
EWAD teacher and a multilingual-generalist CPDP teacher --- is
close to the ideal case; the multilingual sets chart the
boundary conditions.

% =================================================================
\section{Conclusion}
\label{sec:conclusion}

This paper argues that the central question in low-resource
sequence-to-sequence distillation is not simply how much teacher
signal to use, but \emph{when} that signal should be trusted. On
BanSum, standard KD gives only a marginal gain over CE
(\deltaKDvsCE{} R-L), and gradient-alignment probing shows that
\harmfulRatio{} of training samples provide KD gradients that do
not support validation-loss improvement. Common proxies for
reliability, such as teacher confidence and surface ROUGE
agreement, fail to identify these cases reliably, motivating
methods that estimate distillation usefulness more directly.

We introduced two complementary reliability-aware approaches.
CHAD measures sample-level KD utility through validation-gradient
alignment and amortizes that signal with a lightweight gate, while
EWAD+CPDP combines token-level confidence weighting with a
cross-vocabulary geometric constraint from a second teacher. Both
methods improve substantially over standard KD, gaining
\deltaCHADvsKD{} and \deltaECvsKD{} R-L respectively, and both do
so with a \studentParams{}M-parameter student that outperforms a
fine-tuned Qwen-2.5-3B model. The multilingual XL-Sum analysis
further shows that EWAD+CPDP is most useful when teachers provide
complementary information, but less reliable when teachers are
redundant or jointly weak; across \numLangs{} languages it beats
CE on \multiECwinRatio{} with a small, non-significant mean gain
of \multiECoverBase{}.

Overall, the results suggest a practical principle for selective
KD: teacher supervision should be routed by measured reliability,
not assumed quality. Future work should make these reliability
signals cheaper to estimate, extend them beyond summarization, and
study how teacher diversity, calibration, and student capacity
jointly determine when distillation helps rather than hurts.

% =================================================================
\section*{Limitations}
\label{sec:limitations}

\paragraph{Scope and capacity-gap regime.}
Our primary Bangla experiments target a $4{\times}$
teacher--student capacity gap. The principle that
per-sample or per-token KD impact varies and is poorly
predicted by single-feature-family heuristics is language- and
task-agnostic, but magnitudes on other languages and domains
remain to be established. Preliminary experiments applying
reliability-aware ideas to Qwen-2.5 (32B/14B teachers
$\to$ 3B student, smaller relative gap) did not exceed
direct fine-tuning, suggesting gating-based KD provides
value primarily when the student is substantially smaller
than the teacher.

\paragraph{Multilingual significance and sample size.}
The multilingual EWAD+CPDP analysis
(\S\ref{sec:results-multilingual}) covers \numLangs{}
languages with \samplesPerLang{} samples per language under
a single seed. While the per-set and mechanism patterns
are robust to this design, the aggregate
EWAD+CPDP--vs.--Baseline mean difference is not significant
at $n = \numLangs{}$ (paired $t$, $p = \pECvsBase{}$). A
multi-seed multilingual study --- three or more seeds per
(language, method) cell --- would let us separate the
between-seed and between-language components of variance
and produce significance estimates that the present
single-seed design cannot. This is the primary follow-up
we plan.

\paragraph{CHAD is not validated multilingually.}
Our multilingual study carries forward EWAD+CPDP only,
because EWAD+CPDP led on all four Bangla lexical metrics
and because per-language CHAD probe labelling would have
required \numLangs{} separate offline probe stages. We do
not claim CHAD would fail at scale; whether counterfactual
sample-level gating transfers cross-lingually is left open.

\paragraph{Gate quality and probe cost (CHAD).}
The CHAD gate achieves a moderate AUC of \chadAUC{} on held-out
probe samples; a more accurate gate would presumably yield
further improvement. The gradient-alignment label is a
one-step approximation; multi-step or longer-horizon influence
\citep{koh2017understanding} may produce cleaner supervision
at higher cost. The probe itself uses \probeSize{} samples
(4\% of training) and requires one backward pass per probe
sample, amortized across subsequent runs.

\paragraph{Choice of second teacher (EWAD+CPDP).}
CPDP requires a second teacher whose vocabulary differs from
the student's. We used mT5-base XL-Sum as a natural choice for
Bangla but did not systematically explore alternatives. The
inter-teacher distance $d_{T_1 T_2}$ used as a geometric
anchor in \eqref{eq:cpdp-loss} depends on this choice and may
not be optimal.

\paragraph{Semantic-metric protocol.}
The EWAD+CPDP BERTScore-F1 was not computed in the evaluation
run; all other metrics in Table~\ref{tab:main} are directly
comparable across rows.

% =================================================================
\section*{Ethics Statement}

All datasets used in this work (BanSum, XL-Sum subsets) are
publicly available and were used in accordance with their
respective licenses. We did not collect any new data involving
human subjects. To the best of our knowledge the datasets do not
contain personally identifiable information. Bangla news
articles may reflect societal biases present in the source media;
generated summaries can inherit and potentially amplify these
biases, and deployment in user-facing settings should include
appropriate review. All experiments were conducted on
consumer-grade GPUs to keep environmental cost moderate; our
total compute is estimated at approximately 70
GPU-hours. We release code and trained model checkpoints to
support reproducibility. All model and dataset artifacts were used in accordance with their stated licenses, terms of use, and redistribution restrictions. We used AI assistants (Claude) for writing assistance and coding support during the preparation of this work.

% =================================================================
\bibliography{paper}

@inproceedings{abdaoui2020load,
    title = "Load What You Need: Smaller Versions of Multilingual {BERT}",
    author = "Abdaoui, Amine and
      Pradel, Camille and
      Sigel, Gr{\'e}goire",
    editor = "Moosavi, Nafise Sadat and
      Fan, Angela and
      Shwartz, Vered and
      Glava{\v{s}}, Goran and
      Joty, Shafiq and
      Wang, Alex and
      Wolf, Thomas",
    booktitle = "Proceedings of SustaiNLP: Workshop on Simple and Efficient Natural Language Processing",
    month = nov,
    year = "2020",
    address = "Online",
    publisher = "Association for Computational Linguistics",
    url = "https://aclanthology.org/2020.sustainlp-1.16/",
    doi = "10.18653/v1/2020.sustainlp-1.16",
    pages = "119--123"
}

@inproceedings{bhattacharjee2023banglat5,
    title = "{B}angla{NLG} and {B}angla{T}5: Benchmarks and Resources for Evaluating Low-Resource Natural Language Generation in {B}angla",
    author = "Bhattacharjee, Abhik and
      Hasan, Tahmid and
      Ahmad, Wasi Uddin and
      Shahriyar, Rifat",
    editor = "Vlachos, Andreas and
      Augenstein, Isabelle",
    booktitle = "Findings of the Association for Computational Linguistics: EACL 2023",
    month = may,
    year = "2023",
    address = "Dubrovnik, Croatia",
    publisher = "Association for Computational Linguistics",
    url = "https://aclanthology.org/2023.findings-eacl.54/",
    doi = "10.18653/v1/2023.findings-eacl.54",
    pages = "726--735"
}

@inproceedings{fukuda2017efficient,
  title     = {{Efficient Knowledge Distillation from an Ensemble of Teachers}},
  author    = {Takashi Fukuda and Masayuki Suzuki and Gakuto Kurata and Samuel Thomas and Jia Cui and Bhuvana Ramabhadran},
  year      = {2017},
  booktitle = {{Interspeech 2017}},
  pages     = {3697--3701},
  doi       = {10.21437/Interspeech.2017-614},
  issn      = {2958-1796}
}

@inproceedings{hasan2021xlsum,
    title = "{XL}-Sum: Large-Scale Multilingual Abstractive Summarization for 44 Languages",
    author = "Hasan, Tahmid and
      Bhattacharjee, Abhik and
      Islam, Md. Saiful and
      Mubasshir, Kazi and
      Li, Yuan-Fang and
      Kang, Yong-Bin and
      Rahman, M. Sohel and
      Shahriyar, Rifat",
    editor = "Zong, Chengqing and
      Xia, Fei and
      Li, Wenjie and
      Navigli, Roberto",
    booktitle = "Findings of the Association for Computational Linguistics: ACL-IJCNLP 2021",
    month = aug,
    year = "2021",
    address = "Online",
    publisher = "Association for Computational Linguistics",
    url = "https://aclanthology.org/2021.findings-acl.413/",
    doi = "10.18653/v1/2021.findings-acl.413",
    pages = "4693--4703"
}

@misc{hasan2024bansum,
  title         = {{BanSum}: A Dataset for {Bangla} Abstractive Article Summarization with Multiple Sentences},
  author        = {Hasan, Mahmudul and Arean, Abdullah Ibne Hanif and Khan, Md Mosaddek},
  year          = {2024},
  note          = {Dataset publication}
}

@inproceedings{hasib2023bengali,
  author    = {Hasib, Khan and Rahman, Md. Atiqur and Masum, Mustavi Ibne and De Boer, Friso and Azam, Sami and Karim, Asif},
  year      = {2023},
  month     = {11},
  title     = {Bengali News Abstractive Summarization: T5 Transformer and Hybrid Approach},
  booktitle = {2023 International Conference on Digital Image Computing: Techniques and Applications (DICTA)},
  doi       = {10.1109/DICTA60407.2023.00080}
}

@article{hinton2015distilling,
  title   = {Distilling the knowledge in a neural network},
  author  = {Hinton, Geoffrey and Vinyals, Oriol and Dean, Jeff},
  journal = {arXiv preprint arXiv:1503.02531},
  year    = {2015}
}

@article{islam2020hybrid,
  author  = {Islam, Mahimul},
  year    = {2020},
  month   = {11},
  pages   = {27--38},
  title   = {Hybrid Text Summarizer for Bangla Document},
  volume  = {10}
}

@inproceedings{jiao2020tinybert,
    title = "{T}iny{BERT}: Distilling {BERT} for Natural Language Understanding",
    author = "Jiao, Xiaoqi and
      Yin, Yichun and
      Shang, Lifeng and
      Jiang, Xin and
      Chen, Xiao and
      Li, Linlin and
      Wang, Fang and
      Liu, Qun",
    editor = "Cohn, Trevor and
      He, Yulan and
      Liu, Yang",
    booktitle = "Findings of the Association for Computational Linguistics: EMNLP 2020",
    month = nov,
    year = "2020",
    address = "Online",
    publisher = "Association for Computational Linguistics",
    url = "https://aclanthology.org/2020.findings-emnlp.372/",
    doi = "10.18653/v1/2020.findings-emnlp.372",
    pages = "4163--4174"
}

@inproceedings{joshi2020state,
    title = "The State and Fate of Linguistic Diversity and Inclusion in the {NLP} World",
    author = "Joshi, Pratik and
      Santy, Sebastin and
      Budhiraja, Amar and
      Bali, Kalika and
      Choudhury, Monojit",
    editor = "Jurafsky, Dan and
      Chai, Joyce and
      Schluter, Natalie and
      Tetreault, Joel",
    booktitle = "Proceedings of the 58th Annual Meeting of the Association for Computational Linguistics",
    month = jul,
    year = "2020",
    address = "Online",
    publisher = "Association for Computational Linguistics",
    url = "https://aclanthology.org/2020.acl-main.560/",
    doi = "10.18653/v1/2020.acl-main.560",
    pages = "6282--6293"
}

@inproceedings{kim2016sequence,
    title = "Sequence-Level Knowledge Distillation",
    author = "Kim, Yoon and
      Rush, Alexander M.",
    editor = "Su, Jian and
      Duh, Kevin and
      Carreras, Xavier",
    booktitle = "Proceedings of the 2016 Conference on Empirical Methods in Natural Language Processing",
    month = nov,
    year = "2016",
    address = "Austin, Texas",
    publisher = "Association for Computational Linguistics",
    url = "https://aclanthology.org/D16-1139/",
    doi = "10.18653/v1/D16-1139",
    pages = "1317--1327"
}

@InProceedings{koh2017understanding,
  title     = {Understanding Black-box Predictions via Influence Functions},
  author    = {Pang Wei Koh and Percy Liang},
  booktitle = {Proceedings of the 34th International Conference on Machine Learning},
  pages     = {1885--1894},
  year      = {2017},
  editor    = {Precup, Doina and Teh, Yee Whye},
  volume    = {70},
  series    = {Proceedings of Machine Learning Research},
  month     = {06--11 Aug},
  publisher = {PMLR},
  url       = {https://proceedings.mlr.press/v70/koh17a.html}
}

@inproceedings{koo2025switch,
    title = "{SWITCH}: Studying with Teacher for Knowledge Distillation of Large Language Models",
    author = "Koo, Jahyun and
      Hwang, Yerin and
      Kim, Yongil and
      Kang, Taegwan and
      Bae, Hyunkyung and
      Jung, Kyomin",
    editor = "Chiruzzo, Luis and
      Ritter, Alan and
      Wang, Lu",
    booktitle = "Findings of the Association for Computational Linguistics: NAACL 2025",
    month = apr,
    year = "2025",
    address = "Albuquerque, New Mexico",
    publisher = "Association for Computational Linguistics",
    url = "https://aclanthology.org/2025.findings-naacl.206/",
    doi = "10.18653/v1/2025.findings-naacl.206",
    pages = "3733--3746",
    ISBN = "979-8-89176-195-7"
}

@inproceedings{lin2004rouge,
    title = "{ROUGE}: A Package for Automatic Evaluation of Summaries",
    author = "Lin, Chin-Yew",
    booktitle = "Text Summarization Branches Out",
    month = jul,
    year = "2004",
    address = "Barcelona, Spain",
    publisher = "Association for Computational Linguistics",
    url = "https://aclanthology.org/W04-1013/",
    pages = "74--81"
}

@inproceedings{lin2017focal,
  author    = {Lin, Tsung-Yi and Goyal, Priya and Girshick, Ross and He, Kaiming and Dollar, Piotr},
  year      = {2017},
  month     = {10},
  pages     = {2999--3007},
  title     = {Focal Loss for Dense Object Detection},
  booktitle = {Proceedings of the IEEE International Conference on Computer Vision (ICCV)},
  doi       = {10.1109/ICCV.2017.324}
}

@misc{mirzadeh2020improved,
      title         = {Improved Knowledge Distillation via Teacher Assistant},
      author        = {Seyed-Iman Mirzadeh and Mehrdad Farajtabar and Ang Li and Nir Levine and Akihiro Matsukawa and Hassan Ghasemzadeh},
      year          = {2020},
      eprint        = {1902.03393},
      archivePrefix = {arXiv},
      primaryClass  = {cs.LG},
      url           = {https://arxiv.org/abs/1902.03393}
}

@inproceedings{papineni2002bleu,
    title = "{B}leu: a Method for Automatic Evaluation of Machine Translation",
    author = "Papineni, Kishore and
      Roukos, Salim and
      Ward, Todd and
      Zhu, Wei-Jing",
    editor = "Isabelle, Pierre and
      Charniak, Eugene and
      Lin, Dekang",
    booktitle = "Proceedings of the 40th Annual Meeting of the Association for Computational Linguistics",
    month = jul,
    year = "2002",
    address = "Philadelphia, Pennsylvania, USA",
    publisher = "Association for Computational Linguistics",
    url = "https://aclanthology.org/P02-1040/",
    doi = "10.3115/1073083.1073135",
    pages = "311--318"
}

@InProceedings{park2019relational,
  author    = {Park, Wonpyo and Kim, Dongju and Lu, Yan and Cho, Minsu},
  title     = {Relational Knowledge Distillation},
  booktitle = {Proceedings of the IEEE/CVF Conference on Computer Vision and Pattern Recognition (CVPR)},
  month     = {June},
  year      = {2019}
}

@inproceedings{pruthi2020tracin,
 author    = {Pruthi, Garima and Liu, Frederick and Kale, Satyen and Sundararajan, Mukund},
 booktitle = {Advances in Neural Information Processing Systems},
 editor    = {H. Larochelle and M. Ranzato and R. Hadsell and M.F. Balcan and H. Lin},
 pages     = {19920--19930},
 publisher = {Curran Associates, Inc.},
 title     = {Estimating Training Data Influence by Tracing Gradient Descent},
 url       = {https://proceedings.neurips.cc/paper_files/paper/2020/file/e6385d39ec9394f2f3a354d9d2b88eec-Paper.pdf},
 volume    = {33},
 year      = {2020}
}

@misc{sanh2019distilbert,
      title         = {DistilBERT, a distilled version of BERT: smaller, faster, cheaper and lighter},
      author        = {Victor Sanh and Lysandre Debut and Julien Chaumond and Thomas Wolf},
      year          = {2019},
      eprint        = {1910.01108},
      archivePrefix = {arXiv},
      primaryClass  = {cs.CL},
      url           = {https://arxiv.org/abs/1910.01108}
}

@misc{shleifer2020pretrained,
      title         = {Pre-trained Summarization Distillation},
      author        = {Sam Shleifer and Alexander M. Rush},
      year          = {2020},
      eprint        = {2010.13002},
      archivePrefix = {arXiv},
      primaryClass  = {cs.CL},
      url           = {https://arxiv.org/abs/2010.13002}
}

@inproceedings{wen2023f-divergence,
    title = "f-Divergence Minimization for Sequence-Level Knowledge Distillation",
    author = "Wen, Yuqiao and
      Li, Zichao and
      Du, Wenyu and
      Mou, Lili",
    editor = "Rogers, Anna and
      Boyd-Graber, Jordan and
      Okazaki, Naoaki",
    booktitle = "Proceedings of the 61st Annual Meeting of the Association for Computational Linguistics (Volume 1: Long Papers)",
    month = jul,
    year = "2023",
    address = "Toronto, Canada",
    publisher = "Association for Computational Linguistics",
    url = "https://aclanthology.org/2023.acl-long.605/",
    doi = "10.18653/v1/2023.acl-long.605",
    pages = "10817--10834"
}

@article{yang2025qwen3,
  title   = {Qwen3 technical report},
  author  = {Yang, An and Li, Anfeng and Yang, Baosong and Zhang, Beichen and Hui, Binyuan and Zheng, Bo and Yu, Bowen and Gao, Chang and Huang, Chengen and Lv, Chenxu and others},
  journal = {arXiv preprint arXiv:2505.09388},
  year    = {2025}
}

@inproceedings{you2017learning,
  author    = {You, Shan and Xu, Chang and Xu, Chao and Tao, Dacheng},
  title     = {Learning from Multiple Teacher Networks},
  year      = {2017},
  isbn      = {9781450348874},
  publisher = {Association for Computing Machinery},
  address   = {New York, NY, USA},
  url       = {https://doi.org/10.1145/3097983.3098135},
  doi       = {10.1145/3097983.3098135},
  booktitle = {Proceedings of the 23rd ACM SIGKDD International Conference on Knowledge Discovery and Data Mining},
  pages     = {1285--1294},
  numpages  = {10},
  series    = {KDD '17}
}
% (acl.sty already sets \bibliographystyle{acl_natbib})

% =================================================================
\appendix

\section{EWAD+CPDP implementation details}
\label{app:ewad}

This appendix supplements \S\ref{sec:method-ec} with
implementation details not essential to the main exposition.

\paragraph{Numerical stability.}
Both student and teacher logits are cast to float32 before
softmax computation in the EWAD loss \eqref{eq:ewad-loss}. The
teacher is held in bf16 in memory but its logits are upcast
on-the-fly for KL evaluation; this prevents the underflow
observed when computing $\log p^T_t$ from bf16 logits with very
sharp distributions ($p^T_{\max,t} > 0.99$). The CPDP projection
layers are kept in float32 throughout.

\paragraph{CPDP teacher details.}
The vocabulary-incompatible second teacher used by CPDP is
\texttt{csebuetnlp/mt5-base-xlsum-bengali}, a fine-tuned
mT5-base (580M parameters) with a 250{,}112-token SentencePiece
vocabulary. It shares no token ids with the BanglaT5 student;
direct logit-level KL is therefore undefined, motivating the
encoder-hidden-state projection used in CPDP. The mT5 teacher
is run with the same maximum source length as the student
(512 tokens at the 20K stage, 768 tokens at the full-141K
stage).

\paragraph{Projection layer initialization.}
The projection maps $W_S, W_{T_1}, W_{T_2}$ are initialized
with Xavier-uniform scaling. The anchor distance $d_{T_1 T_2}$
in \eqref{eq:cpdp-loss} is computed once after initialization
and held fixed throughout training. We empirically observe that
recomputing $d_{T_1 T_2}$ during training causes the CPDP loss
to drift toward zero by adapting both projections, which removes
the geometric supervision; the detach is necessary.

\paragraph{Training hyperparameters.}
EWAD+CPDP is trained with learning rate $3 \times 10^{-4}$, batch
size 8 with gradient accumulation 4 (effective batch 32), bf16
mixed precision, 8 training epochs maximum with early stopping
patience 2 on validation ROUGE-L, warmup ratio 0.06, gradient
clipping max-norm 1.0. Maximum source length is 512 tokens at
the 20K ablation stage and 768 tokens at the full-141K stage.
Maximum target length is 200 tokens throughout. Both teachers
are loaded in bf16 and held frozen in eval mode. All three
models (student plus two teachers) fit on a single 24 GB GPU.

\paragraph{Relationship to CHAD.}
EWAD+CPDP and CHAD address the same broad question 
when should teacher supervision dominate over gold supervision?
 at different granularities and using different signals.
EWAD+CPDP operates per-token using teacher confidence (EWAD)
and per-sample using inter-teacher geometric divergence (CPDP).
CHAD operates per-sample using counterfactual validation impact.
In Table~\ref{tab:main} on BanSum the two methods perform within
$\deltaECvsCHAD{}$ ROUGE-L of each other (EWAD+CPDP slightly
higher), suggesting both signals capture comparable amounts of
reliability information on this benchmark; we analyse where each
is preferred in \S\ref{sec:analysis}.

\section{Probe analysis details}
\label{app:probe}

\paragraph{Helpful ratio by dataset scale.}
On the 20K filtered subset the gradient-alignment probe labels
80.3\% of \probeSize{} samples as KD-helpful. On the full 141K
dataset the same procedure labels only \helpfulRatio{} as
KD-helpful, with \harmfulRatio{} classified as KD-harmful or
neutral. The shift reflects the broader topic and length
distribution of the full dataset, introducing more ambiguous
articles where the teacher's soft labels are less reliable.

\paragraph{Gate quality.}
Table~\ref{tab:gates} reports held-out AUC for all three gate
configurations on the probe set.

\begin{table}[h]
\centering\small
\begin{tabular}{lccc}
\toprule
\textbf{Gate} & \textbf{Features} & \textbf{AUC} & \textbf{Type} \\
\midrule
A4: entropy  & teacher conf.\ + len  & \entAUC  & logistic \\
A5: ROUGE    & teacher--gold + len   & \semAUC  & logistic \\
CHAD (A6)    & all 13 features       & \chadAUC & GBM regressor \\
\bottomrule
\end{tabular}
\caption{Gate AUC on held-out probe samples. CHAD's AUC of
\chadAUC{} exceeds heuristic gates on both feature subsets.
The downstream ROUGE-L gap between CHAD and the heuristic gates
(Table~\ref{tab:main}) is substantially larger than the AUC gap,
indicating that even modest improvements in gate accuracy have
outsized impact on training dynamics.}
\label{tab:gates}
\end{table}

% -----------------------------------------------------------------
\section{Hyperparameter and training details}
\label{app:hyperparameters}

\begin{table}[h]
\centering\small
\begin{tabular}{lcc}
\toprule
\textbf{Hyperparameter} & \textbf{20K ablation} & \textbf{Full 141K} \\
\midrule
\multicolumn{3}{l}{\textit{Shared (CHAD ablations A1--A6)}} \\
Optimizer          & \multicolumn{2}{c}{AdamW} \\
Learning rate      & \multicolumn{2}{c}{$5\times10^{-5}$} \\
Per-device batch   & \multicolumn{2}{c}{4} \\
Gradient accum.    & \multicolumn{2}{c}{2 (eff.\ batch = 8)} \\
Precision          & \multicolumn{2}{c}{bf16} \\
$\lambda$ (KD wt.) & \multicolumn{2}{c}{0.5} \\
Temperature $\tau$ & 2.0    & 0.5 \\
Max source tokens  & 512    & 768 \\
Max target tokens  & 128    & 200 \\
Beam width         & \multicolumn{2}{c}{4} \\
Epochs             & 13     & 8 \\
Early stop patience& ---    & 5 \\
\midrule
\multicolumn{3}{l}{\textit{CHAD-specific}} \\
Probe size $|\mathcal{P}|$& 1{,}000 & 5{,}000 \\
Val probe $|\mathcal{V}|$ & 200     & 300 \\
Gate type          & logistic & GBM regressor \\
GBM trees / depth  & ---     & 300 / 4 \\
GBM LR / subsample & ---     & 0.05 / 0.8 \\
\midrule
\multicolumn{3}{l}{\textit{EWAD+CPDP-specific}} \\
Learning rate      & \multicolumn{2}{c}{$3\times10^{-4}$} \\
Effective batch    & \multicolumn{2}{c}{32 (bs=8, accum=4)} \\
Warmup ratio       & \multicolumn{2}{c}{0.06} \\
Grad clip norm     & \multicolumn{2}{c}{1.0} \\
EWAD $k$ / $\delta$& \multicolumn{2}{c}{10 / 0.5} \\
EWAD $\eta_\text{CE}$ & \multicolumn{2}{c}{0.3} \\
CPDP $\alpha$      & \multicolumn{2}{c}{0.05} \\
Projection dim     & \multicolumn{2}{c}{256} \\
\bottomrule
\end{tabular}
\caption{Full hyperparameter settings for all experiments.
CHAD and EWAD+CPDP use different learning rates because EWAD+CPDP
trains projection layers jointly with the student; the higher
rate accommodates the newly initialized projection weights while
the lower rate for CHAD matches standard seq2seq fine-tuning.}
\label{tab:hyperparams}
\end{table}

% -----------------------------------------------------------------
\section{Qwen-2.5 Scaling and Decoder-Only Ablation}
\label{app:qwen}

This appendix provides evidence for two claims in
\S\ref{sec:results-qwen}: (i) that fine-tuned Qwen-2.5-3B
achieves ROUGE-L \qwenROUGEL{} on BanSum, and (ii) that
applying EWAD+CPDP with Qwen-2.5 teachers does not outperform
direct fine-tuning of the 3B student.

\paragraph{Qwen-2.5-3B data-scaling results.}
Table~\ref{tab:qwen-scaling} shows Qwen-2.5-3B performance
at two dataset scales. The 141K model (ROUGE-L 0.2335) is the
number cited in the main paper. The monotonic improvement from
20K to 141K confirms the model benefits from additional data
and is not saturated at the smaller scale.

\begin{table}[h]
\centering\small
\begin{tabular}{lcccccc}
\toprule
\textbf{Scale} & \textbf{R-1} & \textbf{R-2} & \textbf{R-L} &
  \textbf{B-4} & \textbf{BS-F1} & \textbf{Sem} \\
\midrule
20K subset  & .2661 & .1241 & .2160 & .0552 & .7389 & .7175 \\
141K full   & .2815 & .1389 & .2335 & .0590 & .7437 & .7364 \\
$\Delta$    & +.015 & +.015 & +.018 & +.004 & +.005 & +.019 \\
\bottomrule
\end{tabular}
\caption{Qwen-2.5-3B fine-tuned on BanSum at two data scales.
Despite scaling to 141K, ROUGE-L 0.2335 remains below all
encoder--decoder reliability-aware students
(CHAD: \chadROUGEL{}, EWAD+CPDP: \ewadcpdpROUGEL{}).}
\label{tab:qwen-scaling}
\end{table}

\paragraph{8-experiment EWAD+CPDP ablation with Qwen-2.5 teachers.}
Table~\ref{tab:qwen-ablation} reports a systematic ablation
using Qwen-2.5-32B and 14B (4-bit NF4) as teachers distilling
into a Qwen-2.5-3B + LoRA student on the filtered 20K subset.
Direct fine-tuning (Baseline, no KD) achieves the highest
ROUGE-L. Single-teacher and fixed-weight dual-teacher
configurations are comparable to the baseline but do not
surpass it. Confidence-only weighting degrades substantially
(ROUGE-L 0.1169). Full EWAD and EWAD+CPDP also underperform
the no-KD baseline.

\begin{table}[h]
\centering\small
\begin{tabular}{lcccccc}
\toprule
\textbf{Experiment} & \textbf{R-1} & \textbf{R-2} & \textbf{R-L} &
  \textbf{B-4} & \textbf{BS-F1} & \textbf{Sem} \\
\midrule
Baseline (no KD)   & \textbf{.2661} & \textbf{.1241} & \textbf{.2160} & \textbf{.0552} & \textbf{.7389} & \textbf{.7175} \\
Single-T 32B       & .2614 & .1210 & .2114 & .0535 & .7364 & .7116 \\
Single-T 14B       & .2640 & .1165 & .2113 & .0523 & .7377 & .7219 \\
Fixed Weights      & .2632 & .1166 & .2104 & .0521 & .7380 & .7220 \\
Confidence Only    & .1529 & .0519 & .1169 & .0239 & .6657 & .6454 \\
Agreement Only     & .2270 & .0917 & .1756 & .0410 & .7181 & .7016 \\
EWAD Full          & .2282 & .0927 & .1767 & .0420 & .7190 & .7031 \\
EWAD+CPDP          & .2246 & .0907 & .1740 & .0407 & .7167 & .7012 \\
\bottomrule
\end{tabular}
\caption{Reliability-aware distillation ablation with Qwen-2.5
teachers (32B + 14B $\to$ 3B + LoRA, 20K BanSum subset).
Direct fine-tuning dominates across all metrics. This contrasts
with our encoder--decoder setting (Table~\ref{tab:main}) where
a 4$\times$ teacher--student capacity gap allows both CHAD and
EWAD+CPDP to substantially improve over the no-KD baseline.
The failure of reliability-aware distillation here is attributed
to the capacity-gap ceiling: when the student already approaches
teacher-level performance, gating-based routing introduces
variance without providing cleaner supervision.}
\label{tab:qwen-ablation}
\end{table}

\end{document}